\documentclass[letterpaper]{article}

\usepackage{natbib,alifeconf}
\usepackage{url,hyperref}
\usepackage{booktabs}
\usepackage{amssymb}
\usepackage{amsmath}
\usepackage{mathtools}
\usepackage{subcaption}
\usepackage{placeins}
\usepackage{afterpage}
\usepackage{xcolor}

\newcommand\blfootnote[1]{%
  \begingroup
  \renewcommand\thefootnote{}\footnotetext{#1}%
  \endgroup
}

\title{Illuminating the Three Dogmas of Reinforcement Learning under Evolutionary Light}

\author{
    Mani Hamidi$^{1}$ \and
    Terrence Deacon$^{2}$ \\
    \mbox{}\\
    $^1$Department of Computer Science, University of T\"ubingen, Germany \\
    $^2$Department of Anthropology, University of California, Berkeley, USA \\
    mani.hamidi@uni-tuebingen.de
}

\begin{document}
\setcounter{secnumdepth}{2}
\interfootnotelinepenalty=10000

\maketitle

\begin{abstract}
Artificial learning systems are graduating from passive learners to increasingly autonomous agents, lending pragmatic urgency to the question of what constitutes agency. Reinforcement learning (RL) offers arguably the most explicit formulation of agent-environment interaction, built on three core tenets: the environment as a Markov decision process, learning as policy optimization, and the agent as a maximizer of scalar reward. Recent work has called to revise these tenets: reconceptualizing learning as adaptation rather than optimization, broadening goals beyond scalar reward, and noting the absence of a formal theory of the agent in a formalism that so heavily emphasizes the environment. We argue that the artificial life community is uniquely positioned to illuminate this critique and concretize an alternative. We draw on open-ended novelty search as a complementary model of adaptation and goal-directed behavior beyond reward optimization, and ground such evolutionary dynamics in thermodynamic theories of origin-of-life and agency, toward a more biologically faithful and formally grounded account of what it is to be an adaptive agent.
\end{abstract}

\section{Introduction}
\label{sec:introduction}

Artificial learning systems\blfootnote{\textcopyright  2026 Mani Hamidi and Terrence Deacon. Published under a Creative Commons Attribution 4.0 International (CC BY 4.0) license.} are rapidly graduating from passive learners operating under strict human supervision to increasingly autonomous agents acting in the world. With this transition, questions about the nature of agency take on pragmatic urgency, demanding that a rich philosophical literature on agency be distilled into formal, scientifically grounded theories that can inform both the engineering of artificial systems and the understanding of biological ones. While recent interest in alignment and safety has drawn the machine learning community's attention to these questions, the artificial life community has---both directly and indirectly---engaged with the problem of agency for decades. Reinforcement learning (RL) offers arguably the most explicit formulation of agent-environment interaction, built on three core tenets\footnote{We prefer \textit{tenet} to \textit{dogma}: it softens the negative connotation the original authors also resisted, and signals that we aim to re-frame these commitments rather than ``shed'' them.}: (T1) the environment formalized as a Markov decision process, (T2) learning as policy optimization on the MDP, and (T3) the agent as a goal-directed system defined by the maximization of scalar reward. \cite{Abel2024-qj} have recently called for a revision of each of these tenets, arguing that (T1) the emphasis on formalizing the environment has come at the expense of a formal theory of the agent, (T2) learning is better understood as adaptation than terminal search, and (T3) scalar reward is insufficient to capture all that we mean by goals. These questions map remarkably well onto areas where the ALife community has deep and longstanding expertise. In this paper, we draw on ideas from open-ended evolution and thermodynamic theories of learning, adaptation, and the origin of life to engage with each.

Our re-framing of RL, and of learning more broadly, is through an evolutionary lens. Although we take the problem of agency (T1) to be at the root of the other two, we begin with T2, where \cite{Abel2024-qj} offer ``adaptation'' as the antidote to an over-reliance on ``search.'' Despite that term's evolutionary connotations, they do not draw on the rich history of evolutionary algorithms that could give it substance, likely because evolution is usually treated as a cross-generational process, unlike RL's within-lifetime learning. To bridge this gap, we draw on theories of Darwinian selection operating in brains within a single lifetime (Section~\ref{neuro-darwinism}).

We then address T2 (Section~\ref{T2}) by recasting adaptation not as the opposite of search but as a different kind of it: open-ended, \emph{objective-free} novelty search rather than objective-driven \emph{terminal} search. We draw on open-ended evolutionary algorithms \citep{Lehman2008-kx, Lehman2011-cn, Brant2017-er} to operationalize this notion through niching and coevolution.

Third, in Section~\ref{T3} we turn the evolutionary perspective on the reward hypothesis and the contention over whether scalar reward can represent multiple objectives. We identify two precise ways its axiomatic foundations \citep{Bowling2023-vl} fail for biological agents, and trace rewards back through homeostatic setpoints to the evolutionary process that grounds them, leading into T1.

Despite championing evolution as the singular remedy for resolving both T2 and T3, in Section~\ref{T1} we argue that an evolutionary explanation alone cannot capture the defining properties of an agent. Our claim is not that there is anything wrong with evolutionary theory itself, but that it rests on preconditions, including the generation of variation through replication, whose origin in turn demands a naturalistic explanation; particularly given how rarely those preconditions are observed to arise spontaneously.
We engage with the homeostasis-versus-reward theories of motivation that also have a history in framing the RL problem to address this point. Pursuing the homeostatic view to its logical conclusion, we point out thermodynamic formalisms for the origin of life and adaptive behavior as the necessary foundation upon which any theory of agency must be built. Beyond theoretical completeness, we sketch its bearing on resource-rational learning and on open-ended evolution in frontier AI (Section~\ref{implications}).

\subsection{Evolution in the Brain} \label{neuro-darwinism}

The brain is not the only organ that learns to form novel associations and solve previously unseen problems from experience. While even single cells can be attributed a kind of learning capacity previously reserved for brains \citep{Gershman2021-bw}, the adaptive immune system epitomizes non-neural learning, rivaling brains in complexity as well as learning proficiency, if not speed.

This is perhaps why the discovery of clonal selection theory of adaptive immunity made such an impression in understanding learning during the lifetime of biological systems more generally. This discovery suggested that evolutionary selection on semi-randomly generated sequences of recombined somatic DNA provides a viable mechanism of learning to recognize and respond to substances (antigens) that were never seen by either the individual itself or any of its ancestors \citep{Jerne1993-rt}.

One of the leading figures in establishing the clonal selection theory of learning by the immune system, Gerald Edelman, was also one of the early advocates of a similar mechanism of learning by the nervous system \citep{Edelman1993-qz}. This research program came to be known as \textbf{neural Darwinism}, and despite the pursuit of complementary ideas in a similar vein \citep{Calvin1998-kd}, its adoption has been limited.

In more recent years, a set of novel hypotheses have been offered that provide a range of hypothetical mechanisms by which Darwinian evolution can operate in the brain. Much of this work is from the efforts of Sz\`athmary and colleagues, and appears under the umbrella of \textbf{Darwinian neurodynamics (DN)} \citep{Fernando2012-fq, Szilagyi2016-qa}. Under the DN framework, the replicating unit that is subject to evolutionary selection can be the neural dynamics of a single or population of neurons, and is not limited to physically replicating units like neurons or synaptic connections themselves. This is most concretely demonstrated in \cite{Czegel2021-qs}, where a population of recurrent networks (reservoir computers) act as information processing nodes for computations that are then read out via the activation of a single neuron. The replication and propagation of these neural patterns to adjacent processing nodes is done according to the ``fitness'' of the solution that it produced.

While \cite{Czegel2021-qs} interpret each of these activation sequences as a solution to a combinatorial problem, they can equally be interpreted as motor sequences, state representations, or even reward functions. The latter is noteworthy, since it makes existing solutions to the problem of the origin of rewards in RL \citep{Singh2010-bk} relevant to learning that occurs within a lifetime rather than over many generations; we will elaborate on this point in Section~\ref{T3}.

Our re-framing is largely conceptual, agnostic to the implementation details by which an evolutionary logic might be realized in neural tissue. The works of Sz\'athmary and colleagues theorize one comprehensive body of such mechanisms, but not the only one. For example, \cite{Dragoi2023-ci} draws on a similar Darwinian framework to suggest that hippocampal networks are pre-configured to generate motifs of neural activity patterns that are subject to selection and modification via experience. Alternatively, \cite{Pezzulo2016-lb} offer a model where the brain simultaneously hosts multiple representations of potential action sequences (``affordances'') that are competing before one ``survives'' and is executed in a continuous sensory-motor loop. And \citet{Bramley2023-ws} argue that although higher cognition is usually thought too goal-directed for evolution's blind search, the growth of a world model is itself blind local variation and selective retention over concepts: a Darwinian search for theories within a single mind. For the purposes of the discussions below, any such neural paradigm that can accommodate replication with variation, along with competition and selection of some subset of these replicators \textit{within the lifetime of an individual}, meets the necessary criteria for our evolutionary re-framing of RL.

\begin{figure*}[!ht]
\centering
\includegraphics[width=0.82\textwidth]{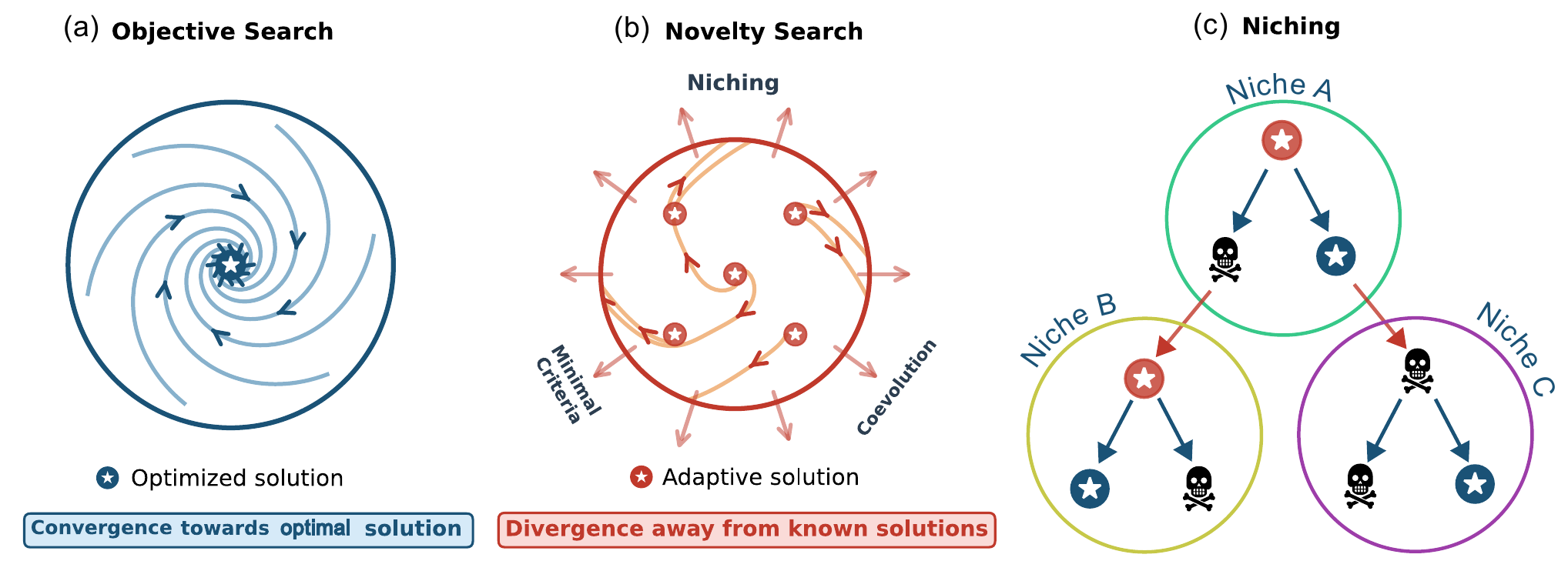}
\caption{Three views of search and adaptation. \textbf{(a)} Objective search as convergent attractor dynamics---trajectories spiral inward toward a terminal solution (``pull''). \textbf{(b)} Open-ended novelty search as divergent repeller dynamics---trajectories push away from visited states, generating diverse adaptive solutions (stars) through three mechanisms: niching, coevolution, and minimal criteria. \textbf{(c)} The resulting speciation into distinct niches, each with its own selection dynamics. Solutions (stars) and failures (skulls) coexist across niches; no single niche dominates.}
\label{fig:adaptation}
\end{figure*}

\section{Adaptation as Novelty Search}
\label{T2}

It is famously said that ``nothing in biology makes sense except in the light of evolution'' \citep{Dobzhansky1973-sa}. Yet evolutionary adaptation is often misconstrued as mere fitness optimization. Simon warns us against this conflation:

\begin{quote}
``...however adaptive the behavior of organisms in learning and choice situations, this adaptiveness falls far short of the ideal of `maximizing' postulated in economic theory. Evidently, organisms adapt well enough to `satisfice'; they do not, in general, `optimize.'\,'' \citep{Simon1956-tg}
\end{quote}

\cite{Abel2024-qj} offer ``adaptation'' as an alternative metaphor to ``search,'' but do not provide a concrete definition of what adaptation entails. Simon's notion of adaptation---satisficing rather than optimizing---operationalized by work in open-ended novelty search \citep{Lehman2011-ms, Pugh2016-qh}, offers exactly such a definition. In this section, we draw on this body of work to articulate what evolutionary adaptation offers to learning that the ``search'' metaphor misses.

The distinction highlighted by \cite{Abel2024-qj} centers on the issue of convergence to a unique solution. This is certainly consistent with the evolutionary view, where many diverse satisficing ``solutions'' to the same ``problem'' (e.g., flight) are often discovered. However, the important insight offered by an evolutionary account is that adaptation is driven by \textit{two complementary engines}: objective-oriented optimization of fitness or ``quality,'' and a concurrent non-objective exploration for novelty or ``diversity'' \citep{Pugh2016-qh, Lehman2011-ms}. The first engine---fitness-driven optimization---is well understood and routinely the sole focus of attention. The second---open-ended diversity generation---is the one that the optimization metaphor misses entirely.

Three mechanisms from the open-ended evolution literature define this second engine. First, \textbf{niching} or \textbf{speciation} \citep{Stanley2002-xw}: organisms found new niches to ``avoid competition and exploit untapped resources'' \citep{Lehman2010-ko}. This principle is recognized far beyond biology. Put bluntly, ``competition is for losers'' \citep{Thiel2014-zero}: the most successful enterprises escape competition by creating new categories rather than outcompeting within existing ones. Second, \textbf{coevolution}: organisms not only occupy niches but actively modify them, resulting in coevolutionary dynamics between agent and environment that further enhance diversity \citep{Odling-Smee2003-gd, Erwin2008-vr}. Such coevolutionary dynamics have been a critical component of open-ended algorithms, where the environment is defined by its own ``genome'' alongside the agent \citep{Brant2017-er, Wang2019-lf}. Third, \textbf{minimal criteria} \citep{Lehman2010-ko, Brant2017-er}: instead of optimizing a fitness function, open-ended algorithms operate on a binary survival threshold---viable or not---which lacks any notion of a gradient or supervision signal. Quality-Diversity (QD) algorithms \citep{Mouret2015-jc, Brant2020-lh} operationalize both engines simultaneously: quality (convergent, within-niche fitness) alongside diversity (divergent, cross-niche expansion). This competitive, diversity-generating logic has direct counterparts in RL: AMIGo's adversarial ``teacher'' invents a curriculum of goals for a student agent \citep{Campero2020-am}, and decomposing such game dynamics into convergent and rotational parts confirms that adversarial coupling contributes divergent, non-convergent flow \citep{Balduzzi2018-nd}.

Figure~\ref{fig:adaptation} contrasts the two engines: a convergent ``pull'' toward a terminal solution versus a divergent ``push'' that speciates into distinct niches with no global ranking across them. Some authors \citep{Dragoi2023-ci} use the dichotomy of ``instructional'' versus ``selectional'' theories to distinguish between these two modes: instructional learning follows a gradient toward a known objective, while selectional learning generates candidates that pre-exist the ``problem'' and are filtered by experience. This is consistent with the ``affordance competition'' framing of \cite{Pezzulo2016-lb}, where the ``affordance landscape'' is constantly adapting not just in reaction to changes in the environment but also the actions of the agent itself. Evolution is not only fitness optimization: non-adaptive processes such as relaxed selection and genetic drift are part of standard evolutionary theory, and appear necessary to explain its rising complexity \citep{Lynch2007-frailty, Deacon2010-yc}.

Echoing \cite{Lehman2025-kb}, we propose that \textit{adaptation}, properly understood, encompasses both engines. The first engine---fitness-driven optimization---excels at solving known unknowns: optimizing a policy when the objective is well-defined and the state space is characterized. The second engine---open-ended novelty search---addresses the unknown unknowns, generating qualitatively new solutions, niches, and even new objectives that were not represented in the original problem formulation. As \cite{Lehman2025-kb} argue, the MDP formalism axiomatically excludes such ``Knightian'' uncertainty, and open-ended evolutionary mechanisms are precisely what equip biological agents to handle it---suggesting that the first engine alone is fundamentally incomplete. Look-ahead planning, gradient-based optimization, and policy improvement remain powerful tools for the known-unknown regime. But adaptation in the full biological sense requires the second engine operating alongside them---generating the diverse building blocks, representations, and objectives that the first engine then exploits.

\section{Beyond the Reward Hypothesis}
\label{T3}

The reward hypothesis---the hypothetical sufficiency of scalar reward maximization in defining all goals---has been axiomatically settled by \cite{Bowling2023-vl}. Their Markov Reward Theorem shows that a Markov reward function consistent with any preference relation exists if and only if five axioms are satisfied: \textit{Completeness}, \textit{Transitivity}, \textit{Independence}, \textit{Continuity}, and \textit{Temporal $\gamma$-Indifference}. We are not contesting the reward hypothesis on its own terms; rather, we use \citet{Bowling2023-vl}'s axiomatic framework to make precise the sense in which the open-ended evolutionary process developed in the previous section is a kind of goal-directed behavior distinct from the objective-driven one the reward hypothesis was designed to capture.

\begin{figure}[!t]
\centering
\includegraphics[width=0.62\columnwidth]{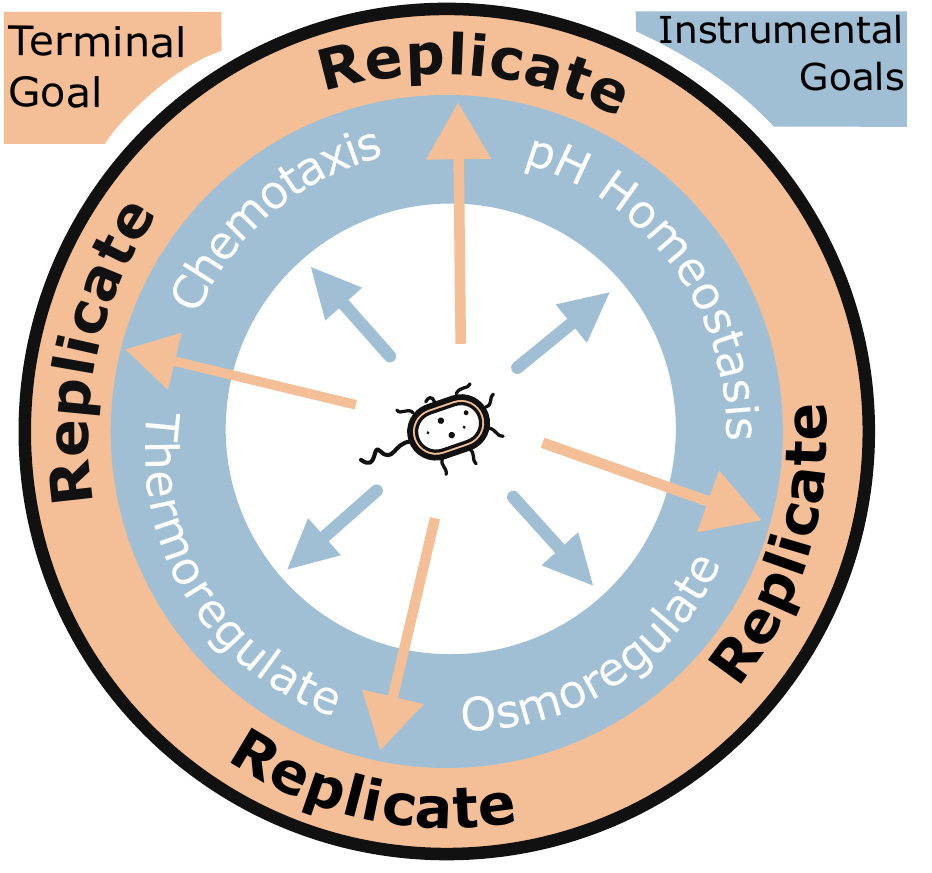}
\caption{A minimal biological agent illustrating the multi-objective structure of biological goals. The outer ring represents the terminal ``goal'' of replication---a non-objective process with no scalar optimum. The inner sectors represent multiple concurrent instrumental goals (chemotaxis, osmoregulation, thermoregulation, pH homeostasis), each homeostatic, each a constraint to satisfy, not a quantity to maximize.}
\label{fig:goals}
\end{figure}

To see why, it is useful to distinguish between two levels of goal-directed behavior (Figure~\ref{fig:goals}). The \textit{terminal} goal of a biological agent is self-persistence: surviving long enough to replicate, thereby sustaining the evolutionary process. The \textit{instrumental} goals are the diverse physiological and behavioral objectives that serve this terminal goal: chemotaxis, osmoregulation, thermoregulation, pH homeostasis, and so on (as schematized for a simple bacterium in Figure~\ref{fig:goals}). In the canonical evolutionary picture, the terminal goal operates across generations while instrumental goals operate within a lifetime. However, as we argued in Section~\ref{sec:introduction}, Darwinian neurodynamics blurs this boundary by enabling evolutionary selection within a single lifetime. The logical distinction between the two levels nevertheless remains, and the reward hypothesis is challenged at each level for different and independent reasons.

\subsection{Completeness Fails: Non-Objective Terminal Goals}

Completeness (Axiom 1) requires that any two outcomes can be compared: for all $A, B \in \Delta(H)$, either $A \succeq B$ or $B \succeq A$. \cite{Bowling2023-vl} note that this excludes the ``incomparability of virtues like `justice' and `mercy''' \citep{Chang2015-zd}. The open-ended evolutionary picture we have outlined produces precisely such incommensurability. As \cite{Lehman2010-ko} illustrate, ``grass does not preclude grasshoppers''---organisms in different ecological niches cannot be ranked on a single preference ordering. Their viability constraints, sensory modalities, and behavioral repertoires occupy non-overlapping regions of phenotype space. There is no fact of the matter about which is ``better''; they are incommensurable. The evolutionary process generates an ever-expanding archive of such incommensurable solutions, none of which dominates the others.

This is the \textit{non-objective} critique: the terminal goal of adaptive agents, namely self-persistence through replication and niche creation, rather than optimization of a fixed objective, is not an optimized objective at all, as it ``falls far short of the ideal of `maximizing' postulated in economic theory'' \citep{Simon1956-tg} after which the reward hypothesis was forged. It is a satisficing criterion that admits no complete preference ordering, and therefore no scalar reward function can represent it.

\subsection{Independence Fails: Instrumental Goals as Minimal Criteria}

At the level of instrumental goals, homeostatic and allostatic theories of motivation \citep{Keramati2014-px, Juechems2019-hk, Verschure2014-iy} have long rivaled the reward-centric view, grounding motivation in multiple concurrent drives with separate setpoints. Multiplicity alone does not defeat the reward hypothesis: a fixed drive function collapses drives into one scalar \citep{Keramati2014-px}, and context-dependent weightings can be absorbed into the state. What resists scalarization is the \emph{form} of a drive. Because physiological variables have lethal limits, a drive functions less as a term to be maximized than as a \emph{minimal criterion} to be satisfied \citep{Brant2020-lh}---the structure that made open-ended search non-objective in Section~\ref{T2}. A goal such as ``gather food while keeping risk within a survival budget'' is a constrained MDP \citep{Altman1999-cm}, and \citet{Bowling2023-vl} show that such goals can violate Independence (Axiom 3) and Continuity (Axiom 4): because feasibility is defined on expectations, mixing in a safe option can make an infeasible strategy feasible and reverse a preference. A constrained goal may admit only a \emph{stochastic} optimal policy \citep{Altman1999-cm, Szepesvari2020-cm}---foraging, say, half the time---whereas every scalar-reward objective admits a deterministic optimum and, being linear in mixtures, satisfies Independence by construction. No scalar reward maximized in expectation can therefore represent such a goal.

Formal and empirical results both support keeping multiple objectives separate rather than collapsing them. Formally, \cite{Miura2022-md} proves that multi-dimensional Markov rewards are strictly more expressive than scalar. Empirically, neuroscience increasingly challenges the scalar assumption: individual dopamine neurons encode heterogeneous, feature-specific prediction errors rather than a uniform scalar signal \citep{Lee2024-fpe}, multiple independent interoceptive reward pathways operate upstream of dopamine \citep{Weber2025-io}, and agents designed with multiple competing ``selves'' outperform monolithic scalar-reward agents in complex, changing environments \citep{Dulberg2022-mod, Dulberg2023-ms}.

\subsection{Where Rewards Come From}

Even within the canonical RL paradigm, setting aside the evolutionary framing we have developed, a more fundamental question remains: where do reward objectives come from in the first place? In artificial intelligence research, goals are construed as reward maximization and the reward function is provided by the engineer. In biological RL, the source of motivation is ultimately attributed to cross-generational evolution \citep{Singh2010-bk}.

Homeostatic theories of motivation \citep{Keramati2014-px, Juechems2019-hk, Verschure2014-iy} offer a more grounded account: reward is derived from physiological feedback control, tracing the origins of goals to the organism's need to maintain its essential variables within viable bounds. These theories are naturally more rooted in biological reality than the abstractions of economic theory where the axiomatic assumptions of the reward hypothesis originate \citep{Morgenstern1964-un}. However, they ultimately defer to evolution for the origins of the setpoints and drive functions themselves.

The question of where rewards come from thus links the multi-objective structure of goals (T3) to the problem of agency (T1). Agency is typically understood in terms of the goals an ``agent'' pursues, and the reward hypothesis is essentially RL's theory of what constitutes such a goal: in this sense the reward hypothesis is also the closest the standard formalism comes to a theory of the agent itself, rather than of the environment in which it acts.
If reward functions are ultimately grounded in homeostatic setpoints, and setpoints are attributed to evolution, then a formal account of goals requires a formal account of the evolutionary process that generates them---and, ultimately, of the thermodynamic conditions that make such a process possible in the first place. We turn to this question next.

\section{Evolution Does Not Explain Agency}
\label{T1}

An RL agent that inherits its reward function from a human engineer can hardly be said to possess agency of its own \citep{Barandiaran2009-ag}; its normativity is borrowed, not generated. Approaches that attempt to remedy this by adding intrinsic motivation objectives such as empowerment or curiosity \citep{Pathak2017-ff, Leibfried2019-tn} do not resolve the problem; they merely replace one human-designed objective with another. Attributing the origins of reward to ``evolution'' \citep{Singh2010-bk}, while more biologically grounded, only passes the explanatory burden one level deeper. The deferral is in fact a regress: standard RL takes reward as given; homeostatic theories ground reward in deviation from physiological setpoints \citep{Keramati2014-px}; setpoints are referred to ``evolution''; and the \textit{open-ended} adaptation we championed for T2 (Section~\ref{T2}) defers in the same direction, since niching, coevolution, and minimal criteria all presuppose a Darwinian process in the first place. Whether we ask where reward comes from (T3) or where open-ended goals come from (T2), the regress runs to the same place: an evolutionary process that does not explain its own preconditions---the survival, replication, and homeostatic dynamics of the very self that its minimal criteria already presuppose.

Evolution cannot be where the regress halts, and one reason can be stated and set aside at once: evolution is non-teleological \citep{dawkins1986blind, gould2002structure}, so it cannot itself be the bearer of the agency we are trying to ground. This was already sensed by the theory's own discoverers: Darwin doubted that the science of his day could conceive a theory for the origins of life \citep{Darwin1863Athenaeum}, while Wallace resorted to homuncular arguments for a \textit{force vitale} \citep{Wallace1912vital}. The substantive work therefore falls to the preconditions themselves, which we approach from two directions.

The first is thermodynamic. Holding an organization away from equilibrium is the problem Schr\"odinger framed as ``negentropy,'' just as heredity was the problem he framed through the ``aperiodic crystal'' \citep{Schrodinger1946-lf}. Thermodynamics has advanced enormously since Darwin and Wallace: the early thought experiments linking it to life, inference, and information (Maxwell's ``demon'' \citep{maxwell1871theory} and Schr\"odinger's two conundrums) have since given way to robust quantitative theories of systems away from equilibrium \citep{jarzynski1997nonequilibrium, Crooks1999-fr}, with applications to the thermodynamics of replication \citep{England2013-eu}, evolutionary adaptation \citep{Perunov2016-fy}, and even sensory-motor control and RL \citep{Hack2023-dt, Grau-Moya2017-gy}. Yet despite a rich literature on the formal correspondence between non-equilibrium statistical physics and inference \citep[e.g., ][]{LaMont2019-yf, Pachter2024-rx, Sohl-Dickstein2015-ga, Zdeborova2015-aq}, its relevance to agency in RL specifically has received much weaker emphasis. This is perhaps because nebulous properties of higher organisms, such as consciousness and sentience, are conflated with the notion of ``agency.'' We instead focus on a more minimal and tractable conception, defined at the dawn of life by a self-persisting, evolvable unit that keeps thermal equilibrium at bay. Section~\ref{negentropy-sub} takes up this thermodynamic precondition directly.

The second direction concerns heredity, and exposes a gap in the very origin-of-life theories meant to supply it. Those theories that envision a self-replicating template molecule, like RNA or DNA, are more favored than the alternatives \citep{Mariscal2019-bm}; this is unsurprising given the success of the ``central dogma of biology'' with DNA as its central player. But a template that arrives already endowed with the properties we wish to explain is a near-miracle, a single improbable, fortuitous event invoked to conjure all the desired properties at once: chance standing in for a mechanism. Even Crick conceded that the origin of life ``appears at the moment to be almost a miracle'' \citep{Crick1981-life}. And even were such a template granted, the emphasis on ``information-carrying'' molecules avoids the problem of agency, offering no account of the system required to read and interpret the purported ``information'' there \citep{Deacon2007-sb, Deacon2008-sb}. Section~\ref{aperiodic-sub} takes up this second puzzle.

Our target throughout is a single minimal feature: \emph{self-persistence through replication}. Its presence would do two things at once. It would earn the word ``self'' in a naturalistic theory of agency by satisfying the requirements for an agent set out by \citet{Barandiaran2009-ag} intrinsically rather than by observer attribution, and by defining the relevant minimum against the second law of thermodynamics rather than against a designer's stipulation. And it would supply the open-ended evolutionary process that we attributed, in the previous sections, to an agent's ultimate goal: a self-persisting, self-replicating unit can sustain the Darwinian dynamics on which Sections~\ref{T2} and~\ref{T3} rely, with the heritable, template-borne variation those open-ended mechanisms require. Theories of the origin of life contend with this same problem, and the best of them yield substrate-independent principles for how an evolutionary process can be spawned, principles that need not remain confined to prebiotic chemistry, as the Darwinian neurodynamics of Sz\'athm\'ary and colleagues (Section~\ref{neuro-darwinism}) already illustrates in a different substrate.

\subsection{Negentropy: The Thermodynamic Precondition}
\label{negentropy-sub}

The two capacities an evolutionary process presupposes, self-maintenance and self-replication, are at bottom a single physical feat: holding an organization away from thermodynamic equilibrium, and reproducing it, against the relentless pull of the second law. To ground evolution is therefore to give a thermodynamic account of these two capacities.

The most direct attempt identifies life with the far-from-equilibrium dissipative structures of Prigogine, such as B\'enard convection, chemical oscillators, and other self-organizing patterns that locally invert the second law. The statistical physics of self-replication \citep{England2013-eu} and of dissipative adaptation \citep{Perunov2016-fy} show, rigorously, that such driven systems are thermodynamically favored to absorb and dissipate work ever more effectively.

There is, however, a fundamental sense in which such structures negate the defining property of a biological agent. Where an agent is occupied with its own self-preservation, a purely dissipative structure is, by construction, self-eliminating: it builds the ordered flows that discharge its sustaining gradient as fast as possible, and so owes its very existence to the rate at which it consumes the conditions of its own persistence. A convection cell or a candle flame races toward the equilibrium that ends it.

The seed of the correction is nonetheless already present in this literature. Not all dissipation is equal: separating the work that drives a system toward an organized, improbable configuration from the work merely squandered in futile cycling distinguishes a campfire from a bacterium, though both are prodigious dissipators \citep{Perunov2016-fy}. But marking this distinction is not the same as generating it. The dissipative-adaptation formalism \emph{accounts for} the difference without supplying the \emph{architecture} that makes dissipation constrained rather than runaway; and where a recent stochastic-thermodynamic model of autonomous agency \citep{Aguilera2024-hy} does build such an architecture in, it does so by imposing a circular constraint, an externally supplied ``charger'' and a geometry in which dissipation is productive by construction, without accounting for where that constraint comes from. What neither supplies is an organization in which a system's dissipative activity is turned back upon preserving the conditions of its own persistence, and which produces that very constraint itself.

The theory of \emph{autogenesis} offers exactly this process-level account \citep{Deacon2014-cr, Deacon2021-fi, Deacon2023-oa, Garcia-Valdecasas2024-by, Deacon2026-au}. Its premise is that two dissipative processes, each self-eliminating in isolation, can be coupled so that each generates the boundary conditions the other requires. In the minimal model (a non-parasitic, ``autogenic'' virus), reciprocal catalysis, which builds up a set of mutually producing catalysts, and molecular self-assembly, which encloses them in a container, share a common product, so that catalysis proceeds fastest precisely where a container forms around it. Neither process can run away: catalysis is contained, and the container is continually rebuilt by catalysis. What emerges is a higher-order, multiply realizable constraint: the \emph{hologenic constraint} \citep{Deacon2021-fi, Deacon2026-au} that belongs not to either chemical process but to their codependence, and that is preserved across cycles of damage and repair.

This constraint is the locus of a minimal, non-homuncular agency: an individuated unit, with an unambiguous self/non-self distinction, that ``acts on its own behalf'' and whose normativity (it can succeed or fail at persisting) is produced rather than borrowed \citep{Garcia-Valdecasas2024-by, Barandiaran2009-ag}.

Against the thermodynamic accounts just surveyed, autogenesis thereby earns what they assume. Where the stochastic-thermodynamic model takes its productive constraint as given, the autogen harvests its free energy through its own catalysis and must \emph{produce} the very constraint that renders its dissipation productive; its work is the continual self-construction of the circular constraint that more abstracted models assume for free. And it unifies into one capacity what those formalisms treat separately: the self-maintaining state whose signature is sustained entropy production \citep{Aguilera2024-hy} is the autogen at rest, while the self-replication whose thermodynamic cost is bounded by the ratio of birth to death rates \citep{England2013-eu} is a special case of the same distributed repair \citep{Deacon2026-au}. This account joins, and can be read as extending, a tradition of autonomy theories in artificial life, including autopoiesis and the closure-of-constraints framework \citep{Maturana-Rumesin1991-in, Montevil2015-xv, Garcia-Valdecasas2024-by}. But its decisive addition is a third one, which carries us out of the negentropy puzzle and into Schr\"odinger's other: the autogen makes this whole capacity \emph{heritable}.

\subsection{The Aperiodic Crystal: From Template to Interpreter}
\label{aperiodic-sub}

Because the hologenic constraint is a relational invariant rather than any particular molecular configuration, it is functionally equivalent to genetic information without being any particular molecule, and can therefore be transmitted across cycles of damage, repair, and reproduction, offloaded over evolutionary time onto dedicated template molecules \citep{Deacon2021-fi, Deacon2026-au}. Heredity, on this account, \emph{emerges from} the self-maintaining system rather than preceding it; the interpreter comes before the template it will eventually read. This is the bridge between Schr\"odinger's two puzzles: the same coupled dynamics that answer the negentropy question give rise, without further miracle, to the aperiodic crystal.

This is why a process-level account closes the gap the template-first story leaves open. Autogenesis is best understood not as a metabolism-first theory but as a ``synergy-first'' fourth paradigm: distinct from replication-first, metabolism-first, and containment-first accounts alike, and alone among them in explaining the origin of molecular information itself \citep{Deacon2026-au}. Its payoff is exactly what an appeal to a fortuitous accident forecloses. A mechanism, unlike a singular lucky event, is the kind of thing that can be \emph{provisioned more than once}, whether engineered or rediscovered by natural selection in some other substrate, turning multiple realizability from a fact about what is possible into a recipe for what can be produced. And a mechanism makes commitments that can fail, where a one-off accident is conducive to neither reproduction nor refutation: chance ``explains'' any outcome equally well, and a singular event cannot be re-run. The synergy-first account is thus not merely the more satisfying story but the more testable one, a point we return to in considering the broader implications of quantifying such a theory (Section~\ref{implications}).

The relational character of the constraint is also what lets the account generalize beyond chemistry. What carries over is not autocatalysis or molecular self-assembly as such, which have no literal counterpart in a nervous system, but the thermodynamic setup itself: two self-undermining dissipative processes pitted into mutual constraint. Stated that abstractly, the schema invites candidate realizations in the neural and physiological realm; one natural place to look is the coupling between metabolic activity and neural signalling, each shaping the conditions of the other \citep{Jacob2023-sq}. Read this way, the autogenic logic is a substrate-independent recipe for spawning an evolutionary process, of which the Darwinian neurodynamics that Sz\'athm\'ary and colleagues locate in neural rather than genetic substrates \citep{Szilagyi2016-qa} (Section~\ref{neuro-darwinism}) would be one instance; notably, one in which a pattern is \emph{self-produced} rather than merely copied.

We do not claim to have delivered that account. We have not shown how an autogenic process actually spawns a Darwinian dynamic in the brain; what we have offered is a template for how one might do so: the mutual-constraint schema above, which awaits its biological candidates. Nor is a minimal, thermodynamic notion of agency (a self-sustaining, replicative process that facilitates evolutionary dynamics) by itself sufficient to conjure a satisfying theory of agency in humans: it does not address the complications of the multicellular transition, let alone of animals with a central nervous system. It is, however, a \textit{necessary} prerequisite for any such theory.

This is the sense in which thermodynamics, and not evolution, finally bottoms out the chain of deferral: each prior link defers its norms to the next, but a self-producing, self-maintaining unit answers to nothing further than the second law, and does so on its own behalf. The picture is less that of a structure than of a \emph{ratchet}: one that, unlike Feynman's \citep{Feynman1963-le, Jarzynski1999-rp}, builds and rebuilds its own pawl, holding fast the organization it has won against the slide back to equilibrium. As for a mathematical formalism of the agent in an RL system, the implication is direct: insofar as the origin of the reward function rests on evolutionary or homeostatic explanations, a formal account of the agent must likewise be grounded in a necessarily embodied agent, one with its proverbial skin in the ultimate game of beating the second law of thermodynamics.

What this perspective brings into focus is the thermodynamically constrained processes that spawn evolutionary dynamics in the first place---whether in a ``warm little pond'' \citep{Darwin1871} or a warm little brain.

\section{Discussion}
\label{implications}

We have viewed three core tenets of RL under evolutionary and thermodynamic light. For T2, open-ended evolutionary theory supplied a precise vocabulary for what \emph{adaptation} means beyond convergence to optimal fitness: two complementary engines, fitness-driven optimization and open-ended diversity generation, operating alongside one another. For T3, we identified two distinct axiomatic failures of the reward hypothesis for adaptive agents: Completeness fails because the equally viable, incommensurable options that evolution generates across niches admit no single ordering,
and Independence fails within an agent juggling multiple objectives, whose preference ordering shifts with context and internal state.
For T1, we argued that the chain of explanatory deferral from reward to homeostasis to evolution bottoms out only in the thermodynamics of self-persistence, self-replication, and ultimately heritable variation on the replicating unit. Grounding evolution is therefore a matter not only of an agent's thermodynamic ``skin in the game,'' but equally of the informational heredity through which its strategies are passed on and varied.

This thermodynamic framing is less of a departure than it may appear: information-theoretic accounts of bounded and resource-rational decision-making already cast an agent's resource limits in thermodynamic terms \citep{Ortega2013-td}, extended even to the non-equilibrium regime of agents adapting in changing environments \citep{Grau-Moya2017-gy}. Where such accounts treat resource-rationality as a cost layered onto an otherwise reward-maximizing agent, the metabolic, second-law-bound grounding we have pursued suggests it is not an appendage to agency but its defining feature.

These ideas are no longer confined to ALife simulations. Open-ended evolutionary principles are increasingly the engine of automated discovery at the frontier of AI: self-modifying agents that keep an archive of variants, where lower-performing ``stepping stones'' later enable breakthroughs \citep{Zhang2025-dgm, Zhang2026-hyperagents, Sarkar2026-eggroll}, are now being turned on scientific discovery itself, with comparable efforts underway in industry. There is something fitting in this. \citet{Abel2024-qj} cast their critique in Kuhn's terms, and in his final writings Kuhn made this picture explicitly evolutionary: rather than converging on a fixed truth, concepts speciate, evolving to fill new semantic niches incommensurable with their ancestors \citep{Kuhn2022-lw}. That the automation of science should reach for evolutionary mechanisms only recognizes what scientific progress may have been all along.

A single thread runs beneath all three tenets: the notion of \emph{information}. Just as RL takes reward, environment, and objective as given, molecular biology's own ``central dogma'' takes the information in DNA and RNA as primary, the template from which life is read out. But a template is inert, and a reward empty, without a system to interpret it; crediting a molecule with ``information'' while leaving that interpreter unexplained only defers the problem \citep{deacon2011incomplete}. The question of where rewards come from is, at bottom, the question of where biological information comes from, and each is usually assumed without provenance. Autogenesis supplies the missing \emph{interpreter} by inverting the usual order: the self-maintaining, self-producing system comes first, and the template is a later affordance onto which its constraints are offloaded \citep{Deacon2021-fi}. Shifting attention from the molecule that carries information to the dynamics that read it is one move that sheds new light on the central dogma of biology and the three dogmas of RL alike, reuniting what Schr\"odinger left apart: negentropy and the aperiodic crystal, self-maintenance and heredity, as two faces of one physical achievement \citep{Schrodinger1946-lf}.

Finally, we may have introduced a fourth, unappreciated dogma of our own: an emphasis on learning within lifetimes at the expense of innate faculties accrued over cross-generational time \citep{Zador2019-em}. Turning attention to the interpreter offers a way out: an organism's form is not read out from a genetic blueprint but emerges from programmed and self-organized flows of information during development \citep{Turing1952-mo, Collinet2021-mo}. Not all of its information is in the genome; much arises in the genome's interplay with the developmental interpreter \citep{Mitchell2025-gc, Adami2000-bc}, the genome being only one part of a two-part code \citep{Grunwald2007}. We hope this view can inform future evo-devo theories of learning \citep{Shuvaev2024-nx, Hamidi2026-gb}.

\subsubsection*{Acknowledgments}
MH thanks the International Max Planck Research School for Intelligent Systems (IMPRS-IS) for their support. MH is supported by the German Federal Ministry of Education and Research (BMBF): T\"ubingen AI Center, FKZ: 01IS18039A. MH is partly funded by the Deutsche Forschungsgemeinschaft (DFG, German Research Foundation) under Germany's Excellence Strategy--EXC2064/1--390727645. We disclose the use of a generative AI assistant (Anthropic's Claude) for copy-editing and language refinement of the manuscript; all scientific content, arguments, and conclusions are the authors' own.

\footnotesize
\bibliography{main}
\bibliographystyle{apalike}

\end{document}